\documentclass[pdflatex,sn-mathphys-num]{sn-jnl}

\usepackage{graphicx}%
\usepackage{multirow}%
\usepackage{amsmath,amssymb,amsfonts}%
\usepackage{amsthm}%
\usepackage{mathrsfs}%
\usepackage[title]{appendix}%
\usepackage{xcolor}%
\usepackage{textcomp}%
\usepackage{manyfoot}%
\usepackage{booktabs}%

\usepackage{algpseudocode}%
\usepackage{listings}%
\usepackage{makecell}
\usepackage[ruled,linesnumbered]{algorithm2e}
\usepackage{caption} 
\usepackage{float}

\theoremstyle{thmstyleone}%
\theoremstyle{thmstyletwo}%

\theoremstyle{thmstylethree}%

\begin{document}

\title[Article Title]{DoPI: Doctor-like Proactive Interrogation LLM for Traditional Chinese Medicine}

\author[1]{\fnm{Zewen} \sur{Sun}}\email{3022244294@tju.edu.cn}
\equalcont{These authors contributed equally to this work.} 

\author[1]{\fnm{Ruoxiang} \sur{Huang}}\email{3022244369@tju.edu.cn}
\equalcont{These authors contributed equally to this work.} 

\author[1]{\fnm{Jiahe} \sur{Feng}}\email{3023001457@tju.edu.cn}

\author[1]{\fnm{Rundong} \sur{Kong}}\email{krd@tju.edu.cn}

\author[1]{\fnm{Yuqian} \sur{Wang}}\email{3022244080@tju.edu.cn}

\author[2]{\fnm{Hengyu} \sur{Liu}}\email{piang.lhy@link.cuhk.edu.hk}

\author[1]{\fnm{Ziqi} \sur{Gong}}\email{ziqi\_victayria@hotmail.com} 

\author[1]{\fnm{Yuyuan} \sur{Qin}}\email{3022244285@tju.edu.cn}

\author*[3]{\fnm{Yingxue} \sur{Wang}}\email{wangyingxue@cetc.com.cn}

\author[1]{\fnm{Yu} \sur{Wang}}\email{wang.yu@tju.edu.cn}

\affil[1]{\orgname{Tianjin University}, \orgaddress{\city{Tianjin}, 300072, \country{China}}} 


\affil[2]{\orgname{The Chinese University of Hong Kong}, \orgaddress{ \city{Hong Kong SAR}, 999077, Asia, \country{China}}} 

\affil[3]{\orgname{China Electronics Technology Group Corporation Limited}, \orgaddress{\city{Beijing}, Asia, \country{China}}} 



\abstract{Enhancing interrogation capabilities in Traditional Chinese Medicine (TCM) diagnosis through multi-turn dialogues and knowledge graphs presents a significant challenge for modern AI systems.  Current Large Language Models (LLMs), despite their advancements, exhibit notable limitations in medical applications, particularly in conducting effective multi-turn dialogues and proactive questioning.  These shortcomings hinder their practical application and effectiveness in simulating real-world medical diagnostic scenarios.  Motivated by the need to address these limitations, we propose DoPI, a novel LLM system specifically designed for the TCM domain.  The DoPI system introduces a collaborative architecture comprising a guidance model and an expert model.  The guidance model is responsible for conducting multi-turn dialogues with patients, dynamically generating questions based on a knowledge graph to efficiently extract critical symptom information.  Simultaneously, the expert model leverages deep TCM expertise to provide final diagnoses and treatment plans.  Furthermore, this study constructs a multi-turn doctor-patient dialogue dataset to simulate real-world interrogation scenarios and proposes an innovative evaluation methodology that assesses model performance without relying on manually collected real-world consultation data. Experimental results demonstrate that the DoPI system achieves an accuracy rate of 84.68\% in interrogation outcomes, significantly enhancing the model's ability to communicate with patients during the diagnostic process while maintaining professional expertise.}




\maketitle

\section{Introduction} \label{sec1}
\begin{figure}[h]
    \centering 
    \includegraphics[width=0.9\textwidth]{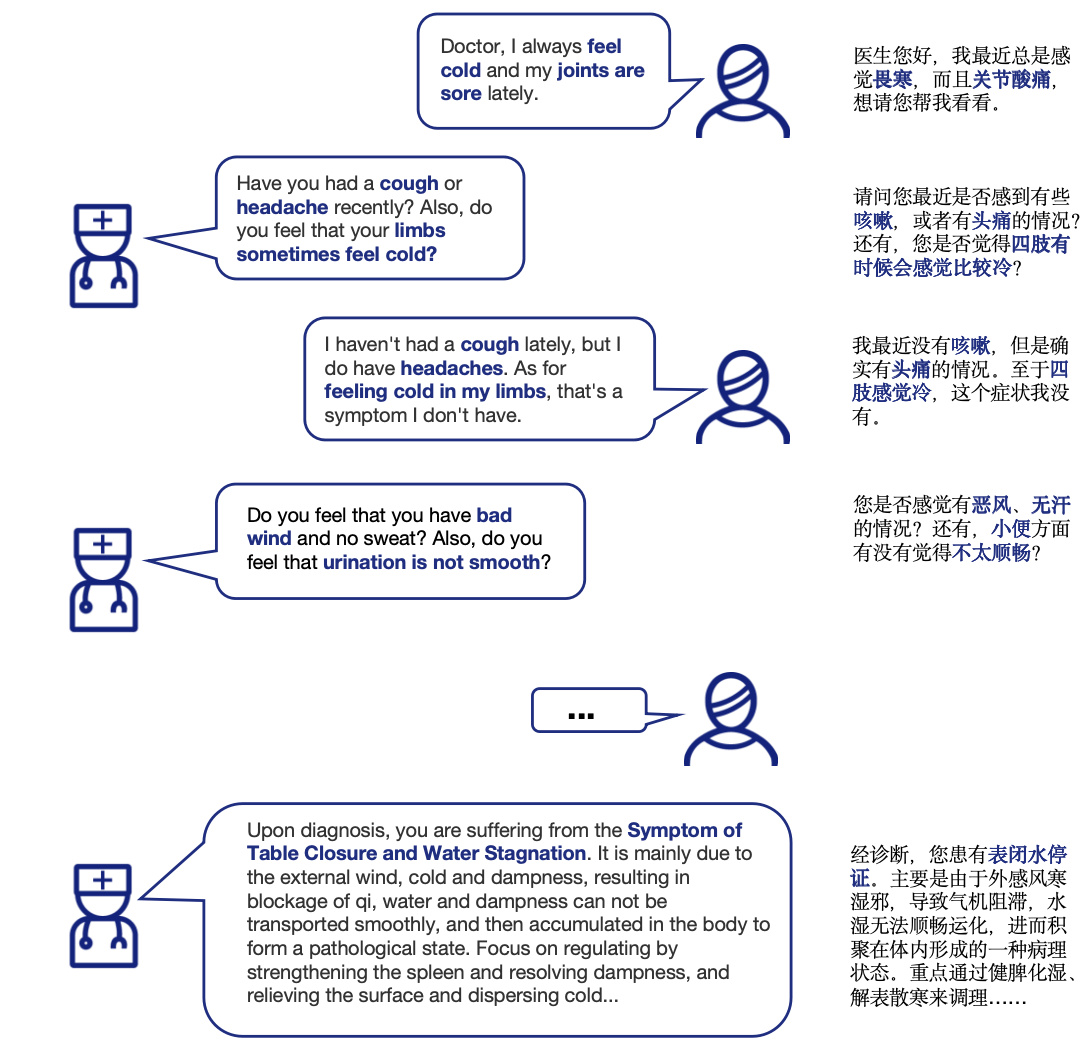}
    \caption{An example of a DoPI having a conversation with a patient demonstrates the DoPI\textquotesingle s proactive interrogation capabilities.} 
    \label{fig1} 
\end{figure}
Large Language Models (LLMs), such as ChatGPT \cite{brown2020language} and LLaMA \cite{touvron2023llama}, have demonstrated profound impacts across various domains due to their advanced capabilities in language comprehension, generation, and knowledge reasoning. However, the generalized nature of these models often limits their effectiveness in addressing domain-specific questions, particularly in fields requiring specialized knowledge, such as medicine. This limitation arises because the training of these models typically involves less specialized domain knowledge and relies on a broad spectrum of underlying data \cite{wang2019edge}.  

Knowledge-enhanced LLMs in the medical domain have been equipped with specialized medical knowledge to perform tasks such as answering medical questions. Currently, several models in Western medicine, including HuatuoGPT \cite{zhang2023huatuogpt}, BianQue \cite{chen2023bianque}, Zhongjing \cite{yang2024zhongjing}, and MedChatZH \cite{tan2023medchatzh}, as well as models in Traditional Chinese Medicine (TCM) such as Sunsimiao \cite{Sunsimiao}, Qibo \cite{zhang2024qibo} and TCM-GPT \cite{yang2024tcmgpt}, have achieved domain knowledge enhancement. This has been accomplished through extensive fine-tuning of model weights or the integration of knowledge graphs via Retrieval-Augmented Generation (RAG) techniques \cite{byambasuren2019preliminary} \cite{he2025opentcm}. Notably, some of these models have even demonstrated the capability to pass the Physician's Licensing Examination.     

\begin{figure}[h]
    \centering
    \setlength{\abovecaptionskip}{3pt}
    \setlength{\belowcaptionskip}{3pt}
    \includegraphics[width=1.0 \linewidth]{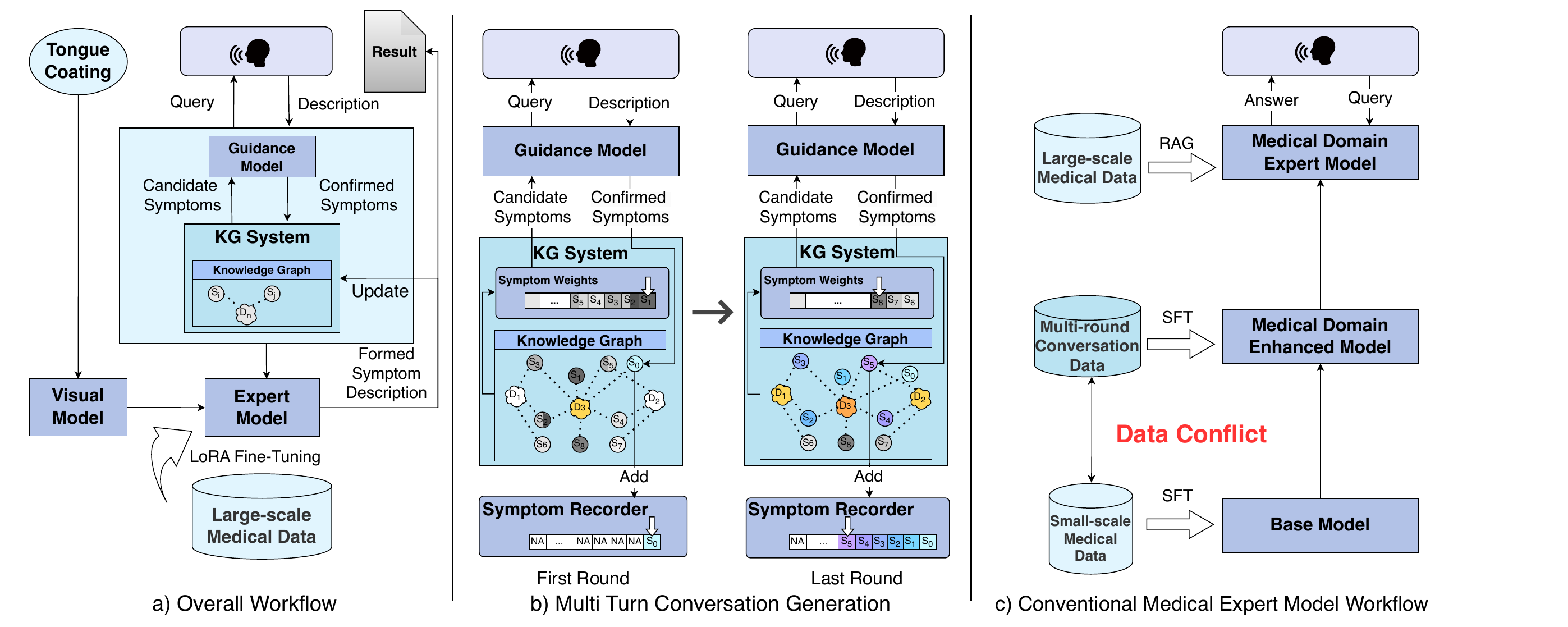}
    \caption{a) illustrates the workflow of the diagnostic and treatment system. The guidance model is responsible for asking questions, while working in conjunction with the Knowledge Graph system to generate proactive multi-turn dialogues. b) illustrates the detailed process of multi-turn dialogue generation: The patient describes their symptoms in natural language to guidance model, which then aligns them with standard medical terminology and submits them to the KG system. The system locates the corresponding symptom nodes (highlighted in color) and adds them to the Symptom Recorder. Grey nodes represent symptoms that have not yet been questioned, with darker shades indicating higher impact factors of symptoms on diseases (the impact factor of the same symptom may vary for different diseases). When asking questions and making judgments, priority is given to symptoms with higher impact factors. c) illustrates the workflow of a conventional medical domain model, where the medical domain dataset was cross-used with the multi-round conversation dataset during training, resulting in data conflicts that impaired the training effect.}
    \label{fig2}
\end{figure} 

However, compared to their performance in knowledge-based question-answering tasks, the capabilities of large models in real-life medical consultation scenarios have declined significantly. The medical large language models mentioned above are limited to providing responses based on the current question and lack the ability to proactively ask patients questions to gather critical information. This is a significant limitation, as many patients are unable to clearly articulate their problems in a single interaction during consultations and often require multiple follow-up questions from doctors to clarify their conditions. The lack of robust multi-turn dialogue capabilities in current medical large models further hinders their ability to effectively complete medical consultations. Moreover, medical consultation does not rely solely on patients' self-reported descriptions. In Western medicine, diagnostic tools such as imaging and laboratory tests often provide more direct insights into a patient's condition. Similarly, in Traditional Chinese Medicine (TCM), tongue diagnosis plays a crucial role in assessing health. Such multimodal data is essential for accurate disease diagnosis. However, current TCM models fail to fully utilize multimodal data, particularly in the case of tongue diagnosis, where the characteristics of tongue coating are highly susceptible to environmental influences, making their integration into models challenging.

We conducted experiments using existing large medical models, including Sunsimiao, HuatuoGPT, and BianQue, on a multi-turn dialogue dataset that we constructed. These experiments revealed significant weaknesses in these models. Specifically, we provided only the initial description of the patient's symptoms as input, which contained partial and incomplete information. Based on the actual symptoms of the sample, we engaged in dialogues with the models and requested them to output a diagnosis of the patient's condition. Our findings indicate that, due to the lack of proactive questioning capabilities, the models often prematurely provided diagnostic conclusions, leading to extremely low accuracy rates.

\begin{table}[h]
\caption{The performance of existing LLMs on dialogue datasets}\label{tab1}
\begin{tabular*}{\textwidth}{@{\extracolsep{\fill}}lccc}
\toprule
\textbf{Metric} & \textbf{Sunsimiao-7B} & \textbf{HuatuoGPT-7B} & \textbf{BianQue-6B} \\
\midrule
Diagnose Accuracy (\%) & 14.73 & 17.72 & 21.27 \\
\bottomrule
\end{tabular*}
\end{table}

The experimental results are shown in Table 1. Sunsimiao, which lacks multi-turn dialogue capability, had a diagnostic accuracy of only \textbf{14.73\%}, while HuatuoGPT and BianQue had diagnostic accuracies of \textbf{17.72\%} and \textbf{21.27\%}, respectively. While HuatuoGPT was capable of engaging in multi-turn dialogues with patients, it rarely asked follow-up questions or sought additional information. On the other hand, BianQue exhibited the ability to ask questions and gather some supplementary information from patients. However, in most cases, the information obtained by BianQue remained insufficient to support an accurate diagnosis of the patient's condition.

Although some large models have been fine-tuned on dialogue datasets and claim to possess multi-turn dialogue capabilities, the so-called multi-turn dialogue process often fails to provide effective information for the final diagnosis. In practice, this process tends to be a superficial dialogue rather than a purposeful and proactive interrogation aimed at gathering critical diagnostic information.

Enhancing the interrogation capabilities of medical LLMs is an urgent task that requires immediate attention. As illustrated in Figure 2(c), most researchers focus on data domain augmentation \cite{cui2023efficient}, employing methods such as Supervised Fine-Tuning (SFT) and Retrieval-Augmented Generation (RAG) to construct medical domain models. In the field of Western medicine, existing approaches enable models to achieve multi-turn dialogue and autonomous questioning capabilities by constructing large-scale doctor-patient dialogue datasets and fine-tuning model weights or applying RAG techniques. However, this approach has a significant drawback: while improving multi-turn dialogue capabilities, it often leads to a reduction in the model's expertise due to changes in model weights or constraints imposed by RAG. This trade-off is not justified by the benefits.In the Traditional Chinese Medicine (TCM) domain, the lack of high-quality TCM dialogue datasets further exacerbates the problem, as current TCM LLMs cannot support specialized TCM questioning. It is evident that fine-tuning LLMs using multi-turn conversation datasets has substantial disadvantages, particularly the loss of model professionalism. Additionally, collecting medical consultation dialogue datasets is highly challenging, and the fine-tuning methods used are not sufficiently generalizable to be effectively migrated to the TCM domain.

To address the aforementioned issues, we propose a novel approach that employs a collaborative framework of two models working in tandem. The guidance model interacts with the expert model through a knowledge graph \cite{byambasuren2019preliminary} to conduct multi-turn dialogues and determines the optimal time to invoke the expert model for providing the final response. We apply this framework to the Traditional Chinese Medicine (TCM) interrogation process to resolve the limitations of current TCM models in conducting effective interrogations. Additionally, we propose an innovative evaluation method that assesses model performance without relying on manually collected real-world consultation data.

As illustrated in Figure \ref{fig1}, DoPI first invokes the medical guidance model to ask the patient a series of questions based on their known symptoms during the consultation process. This enables a more comprehensive understanding of the patient's condition. Once sufficient information is gathered, DoPI calls the expert model to accurately diagnose the patient's disease and provide tailored medical advice.

In summary, the main contributions of this paper are as follows:

$\bullet$ We developed a novel TCM LLM system that integrates an expert model with a guidance model through a knowledge graph. This integration accurately simulates real-world interrogation processes. Without relying on fine-tuning with multi-turn dialogue datasets, the system achieves continuous dialogue capabilities while retaining its professional knowledge.

$\bullet$ We constructed a multi-turn interrogation dataset based on real cases, closely approximating real-world interrogation scenarios and eliminating the challenges of manually collecting such data. We proposed an evaluation method based on the accuracy of interrogation results. Additionally, we extracted a small dataset with low-information validity from the complete dataset to simulate real patient consultation scenarios, enabling the assessment of the model's robustness.

$\bullet$ Our model has achieved an accuracy rate of 84.68$\%$ in the interrogation results on the dataset, significantly improving the ability of communication between model and patients during the interrogation process. 

\section{Backgrounds}\label{sec2}
\subsection{Large Languages Models}

The rapid advancement of large language models (LLMs) has profoundly transformed the field of artificial intelligence. Models such as ChatGPT \cite{brown2020language} and GPT-4 \cite{achiam2023gpt} have achieved qualitative improvements through significant scaling, effectively addressing challenges in natural language processing. The recent introduction of the OpenAI O1 model \cite{zhong2024evaluation} further enhances performance by integrating reinforcement learning with chain-of-thought reasoning, enabling the decomposition and systematic resolution of complex problems. This approach improves both the accuracy and logical coherence of responses. However, OpenAI has not disclosed specific details regarding its training methodologies or weight parameters. As a result, fully open and accessible large models with impressive generative capabilities, such as LLaMA \cite{touvron2023llama}, DeepSeek \cite{liu2024deepseek}, and Bloom \cite{le2023bloom}, have emerged as viable alternatives for research. These models leverage techniques such as reinforcement learning from human feedback (RLHF) and instruction fine-tuning to enhance their performance. Additionally, to improve the quality of Chinese dialogue, their adaptability to the Chinese language has been strengthened through training on extensive Chinese datasets. Significant efforts have also been made to develop proficient Chinese LLMs from the ground up \cite{sun2023moss}\cite{tan2024medchatzh}\cite{hua2024lingdan}\cite{ren2025large}.

\subsection{Fine-tuning LLMs in healthcare}

While LLMs demonstrate remarkable proficiency in general domains, their efficacy in specialized fields, such as healthcare, remains constrained by limitations in training methodologies and scope. In the medical domain, significant progress has been achieved through the development of continuously trained models, including MedAlpaca \cite{han2023medalpaca}, ChatDoctor \cite{chatdoctor}, MedPaLM \cite{singhal2025toward}, and HuatuoGPT \cite{zhang2023huatuogpt}. These models have garnered positive feedback from clinical evaluations and expert assessments. To adapt LLMs to medical contexts, fine-tuning techniques such as LoRA (Low-Rank Adaptation) \cite{hu2021lora} are employed, enabling efficient model adaptation with reduced computational resources. Additionally, these models leverage external specialized medical knowledge through RAG, equipping them with foundational expertise to enhance performance in biomedical tasks.

\subsection{Motivation}
In real-world medical consultations, patients often struggle to articulate all their symptoms clearly in a single interaction. As a result, physicians must engage in proactive questioning to accurately diagnose conditions. However, current LLMs lack this essential capability.

Most existing medical LLMs prioritize knowledge acquisition, constructing datasets to expand their medical knowledge base. While this approach enhances their understanding of medical concepts, it often overlooks the practical aspects of healthcare delivery. Furthermore, existing multi-turn interrogation LLMs are fine-tuned using datasets that include multi-turn dialogues. Although this improves their conversational abilities, the inclusion of low-quality, non-medical content in these dialogues can distort the models' weights, ultimately diminishing their proficiency in medical expertise.

To address these limitations, we propose the development of a novel model that not only retains robust medical knowledge but also incorporates advanced interrogation capabilities. Specifically, our model will be designed to engage in multi-turn, proactive dialogues with patients, simulating real-world diagnostic interactions to more accurately identify and determine diseases.

\section{Model}\label{sec3}

\subsection{Outlook}


As illustrated in Figure 1, we employ a collaborative framework involving a guidance model and an expert model to facilitate multi-turn dialogue interrogation. The guidance model is a LLM fine-tuned on a medical multi-turn interrogation dataset, while the expert model is an LLM fine-tuned on a specialized TCM dataset. These two models operate independently in terms of their knowledge domains but are intricately interconnected throughout the dialogue process. Upon receiving input from the patient, the system evaluates whether a diagnosis can be directly inferred. Based on this assessment, it decides whether to invoke the guidance model to ask further questions or to call the expert model to provide a final diagnosis.

Specifically, when a patient submits a query, the guidance model is initially activated to decompose the input into multiple symptoms. These symptoms are then utilized to identify the most probable diseases within a knowledge graph. Through iterative symptom-based inquiries, the system progressively narrows down the potential diseases to one or a few candidates. Subsequently, the patient's symptoms and the outputs from the guidance model are forwarded to the expert model, which leverages its specialized TCM knowledge to generate a diagnosis and propose a tailored treatment plan.

\subsection{Multi-Round Dialogue Questioning Mechanism Implemented by Knowledge Graphs}

\begin{table}[h]
\centering 
\caption{Notations.}\label{tab2}
\begin{tabular*}{\textwidth}{@{\extracolsep{\fill}}cl} 
    \toprule
    \textbf{Notation} & \textbf{Description}\\
    \midrule
    $G$ & Diseases and Symptoms Knowledge Graph \\ 
    $\epsilon$ & Threshold for diagnosing diseases \\ 
    $Q$ & Patient original query input \\
    $Symptoms$ &The symptoms reported by the patient to the model \\
    $PD$ &The diseases that the patient probably have in the graph \\ 
    $PS$ &The symptoms that the patient probably have in the graph  \\ 
    $PR$ & The response given by the patient to the model's queries about PS \\ 
    $Similarity$ & Current cosine similarity between symptoms and probable diseases \\ 
    $Model\ Answer$ & Model's answer based on symptoms and probable symptoms \\
    \bottomrule 
\end{tabular*}
\end{table}

\begin{algorithm*}[tb]
\caption{The model's interactive diagnostic algorithm.}
\label{alg:MP}
\KwData{Diseases and Symptoms Konwledge Graph $G$, Threshold for dignosing diseases  $\epsilon$, Patient original query input $Q$  }
$Symptoms\ =\  Graph\ Encoder(Q)$\\
\For{$epoch\leftarrow 1\ to\ ask\ epochs$}
{

    \ForEach{$Symptom\ in\ Symptoms\ $}
    {
        $PD\ =\ Find\ Disease\ in\ Graph(G,\ Symptoms)$
        
        $Sort\ According\ to\ Importance(PD)$
        \ForEach{$D\ in\ PD $}
        {
            $PS\ =\ Find\ Symptom\ in\ Graph(G,\ D)$
        }
        
        $Sort\ According\ to\ Importance(PS)$ \\
    }
    $Choose\ some\ Symtoms\ According\ to\ Importance$ \\
    $PR =\ Ask\ Patients(PS)$ \\
    $Update(Symptoms,PR)$ \\
    $Similarity\ =\ Cosine(Symptoms,\ PD)$ \\    
    \If{$Similarity \geq \epsilon $}{
       $Break$
    }    
     
}
$Model\ Answer\ =\ Expert\ Model(Symptoms,\ PD)$\\
$Model\ Update(PS,\ G)$
\end{algorithm*}

We have constructed a TCM knowledge graph as the core of completing multi-turn dialogue consultations with patients. The entire process can be summarized in Algorithm \ref{alg:MP}. Initially, we utilized existing data on TCM disease types and corresponding symptoms to build the TCM knowledge graph. In this graph, symptoms and disease types serve as nodes, while the edges and their weights $w$ represent the degree of association. Initially, the weights are assigned based solely on the frequency of occurrence in the dataset.

For the initial description provided by the user, after parsing through the guidance model, symptoms that can be mapped to the knowledge graph are obtained and recorded as a known symptom group in the Symptom Recorder. Simultaneously, a patient symptom vector P is constructed:
\begin{equation}
P = (p_1, p_2, \ldots, p_n),\quad p_k = \begin{cases} 
1, & \text{If $k$ is a known} \\
    & \text{symptom} \\
0, & \text{otherwise} 
\end{cases} 
\label{P}
\end{equation}
In the questioning loops, several modules are executed iteratively: identifying the next set of symptoms to inquire about, posing questions and parsing user responses, updating the the Symptom Recorder, and determining whether to exit the loop. This process continues until our confidence in the disease differentiation exceeds a predefined threshold.

The process of identifying the next set of symptoms to inquire about relies on the edge weights within the knowledge graph, which represent the significance of relationships between symptoms and between symptoms and diseases. Initially, we traverse all diseases in the graph. For each disease i, we define its symptom vector $D_i$:
\begin{equation}
D_i = (d_{i1}, d_{i2}, \ldots, d_{in}),\quad d_{ik} = w_{ki} \label{D}
\end{equation}
Where $w_{ki}$ represents the edge weight between symptom $k$ and disease $i$. The value of $w_{ki}$ is between 0 and 1. The cosine similarity between the current patient's condition and disease $i$ can be calculated as follows:
\begin{equation}
   S_i = \frac{\mathbf{D_i} \cdot \mathbf{P}}{\|\mathbf{D_i}\| \|\mathbf{P}\|} \label{disease}
\end{equation}
We assign $S_i$ as the importance score for the disease $i$.The higher the score, the more likely the patient is to have disease $i$. Subsequently, we iterate through all unknown symptoms, where $S_i$ can be interpreted as the influence factor of the disease on the symptoms. For each symptom $j$, we calculate its importance score $Score(j)$ by weighting the edge weight $w_{ji}$ between symptom $j$ and disease $i$ with the corresponding influence factor:
\begin{equation}
   Score(j) = \sum_{i=1}^{N} w_{ji} \cdot S_i \label{symptom}
\end{equation}
Notably, to ensure the robustness of symptom discovery, we have incorporated a Gaussian noise perturbation mechanism based on a normal distribution. Specifically, we superimpose a random perturbation term that conforms to an $N(0, \sigma^2)$ distribution on the original importance score calculated for each unknown symptom. Here, the standard deviation $\sigma$ serves as a dynamic adjustment parameter controlling the noise intensity, which we set to decrease as the number of dialogue rounds increases.
\begin{equation}
   FinalScore(j) = Score(j) + \epsilon, \quad \epsilon \sim \mathcal{N}(0, \sigma^2) \label{symptom-f}
\end{equation}
After ranking all unknown symptoms based on their $FinalScore(j)$, we select two or three symptoms with the highest scores as candidate symptoms to inquire about from the patient. Upon receiving the patient's response, we update the Symptom Recorder. This process is facilitated by the guidance model.

Subsequently, we determine whether to exit the questioning loops. When the cosine similarity of the current most probable candidate disease, calculated in Formula \ref{disease}, exceeds a predefined threshold, we conclude that an appropriate disease $D_final$ has been identified. This marks the end of the diagnostic dialogue, and the collected user symptom information from the Symptom Recorder, along with the disease determination $D_final$, are input into the expert model for final diagnosis. The judgment of similarity simulates real-world medical diagnostics, where patients may misjudge the presence or absence of certain symptoms and where disease severity varies. It does not require complete conformity of all symptoms. This setup also introduces an early stopping mechanism to our diagnostic process, preventing unnecessary and redundant dialogue.

It is noteworthy that, inspired by the dynamic path selection and weight adjustment mechanisms in router data transmission, as well as the parameter update characteristics of graph neural networks (GNNs) through chain rule backpropagation \cite{GNN}, we have incorporated an update mechanism into the constructed TCM knowledge graph. Based on the final diagnosis provided by the expert model and the information gathered during the consultation process, the expert model generates update recommendations to either strengthen or weaken the weights between symptoms and between symptoms and diseases within the knowledge graph. Importantly, this update process is executed only after the patient dialogue is concluded, ensuring that it does not introduce any additional delays into the conversation.

\subsection{Medical guidance model}

We utilize a relatively small-scale LLM as our guidance model, which is designed to perform three core tasks:

$\bullet$ Emulating the interaction between a traditional Chinese medicine practitioner and a patient, engaging in multi-turn dialogues to inquire about the presence or absence of relevant symptoms using appropriate natural language.

$\bullet$ Comprehending and parsing the user's responses, facilitating the update of known symptom sets within the questioning loop.

$\bullet$ Aligning medical terminology: when posing questions, accurately conveying information to the user in colloquial language to prevent misunderstandings; when interpreting user inputs, aligning them with symptom terminology stored in the knowledge graph to avoid confusion.

Although two models are utilized throughout the diagnostic process, their invocations are executed serially, with a temporal sequence between them, thereby not increasing the inference cost. Moreover, due to the low performance requirements of the tasks for the guidance model, employing a model of reduced scale effectively decreases the inference cost of the interrogation process.

\subsection{Expert model}

Our expert model leverages the Sunsimiao \cite{Sunsimiao} framework to deliver final diagnoses and treatment recommendations. This model has been fine-tuned on an extensive collection of high-quality TCM datasets, equipping it with profound expertise in TCM. Notably, it has demonstrated exceptional performance in the Chinese National Medical Licensing Examinations for Physicians, Pharmacists, and Nurses. The expert model processes the symptom information and disease differentiations gathered by the guidance model and, drawing on its specialized TCM knowledge, generates accurate diagnostic results and tailored treatment plans.

After concluding the dialogue with the patient, the expert model will propose updates to the symptom information provided by the interrogation model. This includes strengthening the connections between key symptoms and diseases, as well as between key symptoms themselves, while weakening the links between misleading symptoms and diseases, and between misleading symptoms and other symptoms.

Specifically, in addition to processing information from patient dialogues during the interrogation stage, we have integrated tongue diagnosis as auxiliary input. Patients can submit photographs of their tongue coating, which serve as supplementary data. Tongue diagnosis, a critical technique in TCM, evaluates health conditions and disease characteristics by analyzing the color, shape, and features of the tongue coating. However, due to potential inaccuracies in tongue coating recognition and the influence of incidental factors, we avoid invoking a multimodal large model to process the tongue coating as a visual modality. Instead, we employ a ResNet-based convolutional neural network (CNN) to classify the patient's constitution type based on the tongue coating image. This classification acts as a supplementary input to the text modality, combining with symptom information to provide the expert model with a more comprehensive understanding of the patient's condition. This approach enhances the accuracy and effectiveness of diagnosis and treatment.
\section{Construction of Multi-Round Conversation}\label{sec4}

Due to the limited availability of dialogue data, we constructed a multi-turn doctor-patient dialogue dataset by leveraging classical TCM knowledge. This process relies on the capabilities of LLMs. Specifically, we assigned the roles of patient and doctor to the LLMs, a methodology that has proven effective in prior studies \cite{zhang2023huatuogpt} \cite{chatdoctor}.

Initially, we organized the data into multiple binary tuples of 'disease $+$ symptom list' as fundamental medical knowledge units. For each tuple, we assumed that the patient suffers exclusively from the specified disease and exhibits all symptoms listed in the symptom list. Building on this assumption, we utilized the qwen-plus model \cite{qwen25} to simulate and construct the doctor-patient dialogue.

The generated dialogues encompass the entire medical consultation process, including the patient describing symptoms, the doctor inquiring and the patient responding in an iterative loop, and the doctor delivering a final diagnosis. This simulates a comprehensive real-world clinical scenario. Crucially, before each inquiry, the doctor's questioning is guided by calculations based on known information in the knowledge graph, rather than being randomly selected. This approach effectively emulates the logical reasoning process of a TCM practitioner, grounded in relevant medical knowledge.

To ensure the robustness of natural language during the dialogue process, we defined several basic requirements for the patient and doctor large language models:

$\bullet$ Common Sense: To better align with real-world scenarios, the patient's initial description should focus on major, easily observable symptoms. The dialogue should remain objective and consistent with the patient's role.

$\bullet$ Colloquialism: We assume that the patient lacks extensive professional medical knowledge. Therefore, the doctor should use easily understandable language when asking questions, and the patient's responses should be limited to symptom-related information without irrelevant details.

$\bullet$ Honesty: We assume that the patient can accurately judge the symptoms inquired by the doctor and will not report any non-existent symptoms. This ensures that the doctor receives correct information, reducing uncertainty in the multi-turn dialogue dataset.
\section{Evaluations}\label{sec5}
\subsection{Evaluation Methodology}

\textbf{Platforms and Implementations}: 
The experiments were conducted on two NVIDIA A6000 GPUs with a total of 96GB VRAM. Except for ChatGPT, DeepSeek and Qwen, for which we utilized their respective APIs, all other models in the experiments were deployed using the Hugging Face Transformers framework.

\textbf{Datasets, and Baseline}: We used the previously mentioned dataset of multiple rounds of doctor-patient conversations containing more than two thousand pieces of high-quality data. In our experiments, we used only initial descriptions containing only some of the patient's symptoms as input, and the medical model's diagnosis of the patient's disease as output through an automated testing program.

We used the following model as a baseline for our experiments:

$\bullet$ Qwen2.5: Qwen2.5\cite{qwen25}(Qwen2.5-Max) is a generalized Chinese language model that supports long contexts and multiple rounds of conversation

$\bullet$ Sunsimiao: Sunsimiao\cite{Sunsimiao} is a Chinese language model for the medical domain without multi-round conversations.

$\bullet$ HuatuoGPT: HuatuoGPT\cite{zhang2023huatuogpt} is a Chinese language model for instruction fine-tuning using a medical domain knowledge dataset. Since the model weights are not yet open source, we tested them manually on the web side.

$\bullet$ BianQue: BianQue\cite{chen2023bianque} is a Chinese medical domain model focusing on enhanced interrogation capabilities

$\bullet$ ChatGPT: ChatGPT\cite{achiam2023gpt}(ChatGPT-4o) is one of the most popular multimodal LLMs with excellent multi-round conversation and logical reasoning capabilities

$\bullet$ DeepSeek: DeepSeek\cite{liu2024deepseek}(DeepSeek-v3) is an LLM based on the MOE architecture with excellent efficiency and performance compared to other mainstream models.

\subsection{Evaluation with metrics}
We quantitatively assessed the model's proactive interrogation capabilities using the following metrics:

Firstly, to measure the effectiveness of medical modeling for diagnosis, we define the metric of \textbf {Diagnostic Accuracy}:
\begin{equation}
   Diagnostic \ Accuracy = \frac{P_c}{P} \label{Diagnostic Accuracy}
\end{equation}
where $P$ denotes the number of all patients and $P_c$ denotes the number of patients whose disease finally diagnosed by the model is consistent with the disease suffered by the patient.

Next, To measure the proactivity of the model in the questioning process, we define the metric \textbf{Q$\&$A Ratio}:
\begin{equation}
   Q\&A  \ Ratio = \frac{1}{n} \sum^{n}_{i=1} \frac{Q^{m}_i}{A^{m}_i} \label{QA Ratio}
\end{equation}
where $Q^{m}$ denotes the number of rounds in which the model asks the patient a question during the consultation,  $A^{m}_i$ denotes the number of rounds in which the model answers (or makes a diagnosis) to the patient during the consultation and n is the number of patients in the test set.

In addition, to compare the difference between the model and the professional doctor's questioning proactivity during the interrogation process, we defined the metric of \textbf{Interrogation Distance}.
\begin{equation}
   \begin{split}
        Interrogation \ Distance = \frac{1}{n} \sum^{n}_{i=1} \left( (Q^{m}_i-Q^{d}_i)^2 \right. \\
        \left. + (A^{m}_i-A^{d}_i)^2 \right)
    \end{split}  \label{Interrogation Distance}
\end{equation}
where $Q^{m}$ and $A^{m}$ denote the number of rounds in which the model questioned and answered the patient during the interrogation process, $Q^{d}$ and $A^{d}$ denote the number of rounds in which the specialized physician questioned and answered the patient in the dataset, and n is the number of patients in the test set.

\begin{table*}[h]
\caption{Comparison of DoPI and large-scale LLM on interrogation metrics }
\centering
\renewcommand{\arraystretch}{1.5}
\begin{tabular}{cccc}
\Xhline{1.2px}
\textbf{Model}         & \textbf{Diagnose Accuracy} & \textbf{Q\&A Ratio} & \textbf{Interrogation Distance}  \\ \hline
Qwen2.5-Max & 32.31      & 15.23    &2.32  \\
ChatGPT-4o & 35.12    &17.09    & 2.04    \\
DeepSeek-v3 & 58.74    & 20.98   & 1.96  \\
\textbf{DoPI-7B} & \textbf{84.68}     & \textbf{21.31}        & \textbf{1.84}         \\ 
\Xhline{1.2px}
\end{tabular}
\end{table*}

The experimental results are shown in Table 3, and DoPI achieves outstanding performance in several evaluation metrics. By leveraging a robust medical guidance model, DoPI surpasses other models in both the proactivity of questioning and the alignment of its questioning process with that of a professional physician. This alignment contributes to significantly higher diagnostic accuracy, enabling more effective diagnosis and the provision of precise medical advice. 

In comparison with existing medical models of the same size, DoPI has a stronger ability to have multi-round conversations, and can obtain more information about the patient through proactive interrogation, so as to more accurately make a diagnosis of the patient and provide effective advice with a comparable amount of existing medical knowledge. 

\subsection{Evaluation with LLM}

LLMs have been demonstrated to be effective natural language evaluators, and their application in comparative text evaluation can reliably reflect the quality of generated texts \cite{llme}.

In our evaluation, we provided the LLM with two sets of doctor-patient dialogues and evaluation metrics, instructing it to assess which doctor performed better based on the specified criteria and to elucidate the rationale behind its judgment. In instances where a superior performer was discernible, the LLM was tasked with identifying the winner; conversely, if the disparity between the two was negligible, it was to declare a tie. Acknowledging the stochastic nature of LLMs, each comparative assessment was conducted five times, with the most frequently occurring judgment being adopted as the definitive result.

The evaluation encompasses four key criteria:

$\bullet$Knowledgeability: This refers to the volume of medical knowledge demonstrated in the dialogue, including the ability to answer symptom-related questions and make disease judgments.

$\bullet$Professionalism: This assesses whether the diagnostic advice provided is both professional and comprehensive.

$\bullet$Fluency: This evaluates the naturalness and smoothness of the dialogue, ensuring it closely resembles a real-world consultation scenario.

$\bullet$Respectfulness: This measures the extent to which the dialogue shows adequate respect for privacy and other sensitive issues, as well as whether the language used conveys sufficient concern for the patient.


DoPI also demonstrates advantages in comparison with large-scale LLMs. While these large-scale LLMs excel in multi-turn dialogue and logical reasoning capabilities, their questioning during the interrogation process is not grounded in pre-existing medical knowledge. Consequently, they often fail to extract meaningful diagnostic information through interrogation. In contrast, DoPI establishes structured rules based on the relationships between symptoms and the connections between symptoms and diseases. By formulating questions according to these rules, DoPI is more likely to gather additional patient-specific information, thereby enabling more accurate diagnoses.

\begin{figure}[h]
    \centering 
    \includegraphics[width=0.9\textwidth]{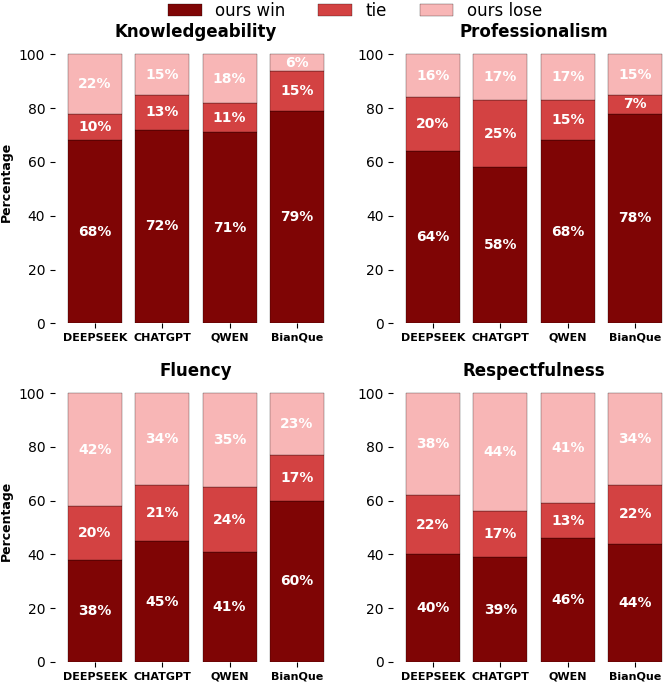} 
    \caption{Evaluation results provided by LLM} 
    \label{fig3} 
\end{figure}

We employed this methodology to conduct comparative evaluations with four models: DeepSeek-v3, ChatGPT--4o, Qwen2.5-max, and BianQue. The evaluation results are illustrated in Figure \ref{fig3}. Our model outperforms the baseline model across the majority of the four evaluation criteria, particularly excelling in knowledgeability and professionalism. Equipped with an extensive repository of medical knowledge and robust multi-turn dialogue capabilities, our model achieves a seamless integration of these features through the knowledge graph we constructed. This integration has enabled the development of a highly effective diagnostic model, offering valuable insights and a solid foundation for future research and practical applications in this domain.

\section{Conclusion}\label{sec6}
The DoPI system significantly enhances Traditional Chinese Medicine interrogation capabilities through the integration of a guidance model and an expert model. Leveraging a knowledge graph, it dynamically generates contextually relevant questions, thereby improving diagnostic accuracy. The system demonstrates superior performance compared to existing LLMs in both diagnostic precision and proactivity during multi-turn dialogues.

\textbf{Data Availability}  The data from the original dataset is not publicly available to protect the personal privacy of patients.

\bibliography{sn-bibliography}

\end{document}